\documentclass[runningheads,a4paper]{llncs}

\usepackage{amssymb,enumerate}
\usepackage{graphicx}

\usepackage{url}
\urldef{\mailsa}\path|{alfred.hofmann, ursula.barth, ingrid.haas, frank.holzwarth,|
\urldef{\mailsb}\path|anna.kramer, leonie.kunz, christine.reiss, nicole.sator,|
\urldef{\mailsc}\path|erika.siebert-cole, peter.strasser, lncs}@springer.com|    
\newcommand{\keywords}[1]{\par\addvspace\baselineskip
\noindent\keywordname\enspace\ignorespaces#1}

\begin{document}

\mainmatter  % start of an individual contribution

\title{Hybrid Approach for Inductive Semi Supervised Learning using Label Propagation and Support Vector Machine}

\titlerunning{Hybrid Approach for Inductive Semi Supervised Learning}

%\titlerunning{Lecture Notes in Computer Science: Authors' Instructions}

\author{Aruna Govada, Pravin Joshi, Sahil Mittal, Sanjay K Sahay}

\institute{BITS, Pilani, K.K. Birla Goa Campus, Zuarinagar-403726, Goa, India}


\toctitle{Hybrid Approach for Inductive Semi supervised learning using Label Propagation and Support Vector Machine}
%\tocauthor{Authors' Instructions}
\maketitle

\begin{abstract}
Semi supervised learning methods have gained importance in today's world because of large expenses and time involved in labeling the unlabeled data by human experts. The proposed hybrid approach uses SVM and Label Propagation to label the unlabeled data. In the process, at each step SVM is trained to minimize the error and thus improve the prediction quality. Experiments are conducted by using SVM and logistic regression(Logreg). Results prove that SVM performs tremendously better than Logreg. The approach is tested using 12 datasets of different sizes ranging from the order of 1000s to the order of 10000s. Results show that the proposed approach outperforms Label Propagation by a large margin with F-measure of almost twice on average. The parallel version of the proposed approach is also designed and implemented, the analysis shows that the training time decreases significantly when parallel version is used.
\end{abstract}

\section{Introduction}

Semi supervised learning methods are mainly classified into two broad classes: Inductive and Transductive. In both Inductive and transductive the learner has both labeled training data set $(x_i,y_i)_{i= 1...l} \sim p(x,y)$ and unlabeled training data set $(x_i)_{i=l+1...l+u} ~ p(x)$, where $l \ll u$. The inductive learner learns a predictor f : X $\rightarrow$ Y , f $\in$ F where F is the hypothesis space, $x \in X$ is an input instance, $y \in Y$ its class label. The predictor learns in such a way that it predicts the future test data better than the predictor learned from the labeled data alone. In transductive learning, it is expected to predict the unlabeled data  $(x_i)_{i=l+1...l+u}$ without  any expectations of generalizing the model  to future test data. Gaussian processes, transductive SVM and graph-based methods fall in the latter category. On the other hand, the former  models are based on joint distribution and examples include Expectation Maximization.
In many real world scenarios, it is easy to collect a large amount of unlabeled data \{x\}. For example, the catalogue of celestial objects can be obtained from sky surveys, Geo spatial data can be received from satellites, the documents can be  browsed through the web. However, their corresponding labels \{y\} for the prediction, such as classification of the galaxies, prediction of the climate conditions, categories of documents often requires expensive laboratory experiments, human expertise and a lot of time. This labeling obstruction results in an insufficiency in labeled data with an excess of unlabeled data left over. Utilizing this unlabeled data along with the limited labeled data in constructing the generalized predictive models is desirable. Semi supervised learning model can either be built by first training the model on unlabeled data and using labeled data to induce class labels or vice versa.
The proposed inductive approach labels the unlabeled data using a hybrid model which involves both Label Propagation and SVM. At every step in the process, it fits the model to minimize error and thus improve the prediction quality. 
The rest of the paper is organized as follows. The related work is discussed in section 2.Label propagation, SVM are discussed in section 3 and the proposed approach is presented in section 4. Section 5 contains the experimental results and comparison of the proposed approach with the Label Propagation algorithm followed by Conclusion and future work in section 6.

\section{Related Work}

A decent amount of work has been done in the field of semi-supervised learning in which major role has been played by the unlabeled data which is in huge amount as compared to the labeled data.Castelli et al.[1],[2] and Ratsaby et al.[3] showed that unlabeled data can predict better if the model assumption is correct. But if the model assumption is wrong, unlabeled data may actually hurt accuracy.Cozman et al.[4] provide theoretical analysis of deterioration in performance with an increase in unlabeled data and argue that bias is adversely affected in such situations. Another technique that can be used to get the model correct is to down weight the unlabeled data by  Corduneanu et al.[5].Callison-Burch et al [6] used the down-weighing scheme to estimate word alignment for machine translation.

A lot of algorithms have been designed to make use of abundant unlabeled data. Nigam et al.[7] apply the Expectation Maximization[8] algorithm on mixture of multinomial for the task of text classification and showed that the resulting classifiers predict better than classifier trained only on labeled data.

Clustering has also been employed over the years to make use of unlabeled data along with the labeled data.The dataset is clustered and then each cluster is labeled with the help of labeled data.Demiriz et al.[9] and Dara et al.[10] used this cluster and label approach successfully to increase prediction performance. 

A commonly used technique for semi-supervised learning is self-training. In this, a classifier is initially trained with the small quantity of labeled data. The classifier is then used to classify the unlabeled data and unlabeled points which are classified with most confidence are added to the training set. The classifier is re-trained and the procedure is repeated. Word sense disambiguation is successfully achieved by Yarowsky et al.[11] using self-training. Subjective nouns are identified by Riloff et.al[12]. Parsing and machine translation is also done with the help of self-training methods as shown by Rosenberg et al.[13] in detection of object systems from images.

A method which sits apart from all already mentioned methods in the field of semi supervised learning is Co-training. Co-training [14] assumes that (i) features can be split into two sets; (ii) Each sub-feature set is sufficient to train a good classifier; (iii) the two sets are conditionally independent given the class. Balcan et al.[15] show that co-training can be quite effective and that in the extreme case only one labeled point is needed to learn the classifier.

A very effective way to combine labeled data with unlabeled data is described by Xiaojin Zhu et al.[16] which propagated labels from labeled data points to unlabeled ones. An approach based on a linear neighborhood model is discussed by Fei Wang et al.[17] which can propagate the labels from the labeled points to the whole data set using these linear neighborhoods with sufficient smoothness.
Graph based method is proposed in [18] wherein vertices represent the labeled and unlabeled records and edge weights denote similarity between them. Their extensive work involves using the label propagation to label the unlabeled data, role of active learning in choosing labeled data, using hyper parameter learning to learn good graphs and handling scalability using harmonic mixtures.

\section{Label Propagation} 

Let $(x_1, y_1)......(x_l, y_l)$ be labeled data, where $(x_1.......x_l)$ are the instances, Y  = $(y_1.....y_l) \in (1....C)$ are the corresponding class labels. Let $(x_{l+1}, y_{l+1}).....(x_{l+u}, y_{l+u})$ be unlabeled data where $Y_U = (y_{l+1}.....y_u)$  are unobserved. $Y_U$ has to be estimated using $X$ and $Y_L$, where $X = (x_1......x_l........x_{l+u})$  \\
$$ F:L \cup U \longrightarrow R$$
$w_{ij}$ is similarity between i and j, where $i,j \in X$
F should minimize the energy function\\
$$(f) =\frac{1}{2}\sum_{i,j} w_{ij}(f(i) - f(j))^2 = f^T\bigtriangleup f$$
and $f_i, F_j$ should be similar for a high $w_{ij}$.

Label propagation assumes that the number of class labels are known, and all classes present in the labeled data.

\subsection{Support Vector Machine}
 The learning task in binary SVM can be represented as the following
$$min_w = \frac{\parallel w \parallel^2}{2}$$

subject to $y_i(w.x_i + b) \ge 1, \quad i =1,2,....N$  where $w$ and $b$ are the parameters of the model for total $N$ number of instances.\\

 Using Legrange multiplier method the following equation to be solved,

              $$ L_p = \frac{\parallel w \parallel^2}{2} - \sum_{i=1...N} \lambda_i (y_i(w.x_i   + b) -1),  \lambda_i$$   are called legrange multipliers. By solving the following partial derivatives, we will be able to derive the decision boundary of the SVM.

   		$$ \frac{\partial p}{\partial w} = 0 \le w \Rightarrow  \sum_{i=1....N} \lambda_i y_i x_i  = 0$$ 
		$$ \frac{\partial p}{\partial b} = 0 \le w \Rightarrow  \sum_{i=1....N} \lambda_i y_i = 0$$

\section{Proposed Approach} 

The proposed Algorithm is an inductive semi supervised learning, an iterative approach in which at each iteration the following steps are executed. In the first step label propagation is run on the training data and the probability matrix is computed for the unlabeled data. The second step is to train the SVM on the available labeled data. Now in the third step, the classes for unlabeled data are predicted using SVM and class probabilities computed in step 1 are compared with the threshold. In step 4, all the records for which both label propagation and SVM agree on the class label, are labeled with corresponding class.This continues till all the records are labeled or no new records are labeled in an iteration. One-one multi class SVM is used as the data consists of multi classes.

The records of each of the data sets are shuffled, 70\% is considered for training and the rest is considered for testing. 80\% of the training data is unlabeled. 

The algorithm is implemented in both the serial and parallel versions.

\subsection {Serial Version}
Input: Classifier, Threshold\\
Output: F-measure

\begin{enumerate}[1.]
\item (labeled\_records, unlabeled\_records) = select\_next\_train\_folds()\\
\# \textit {Each fold of data is split into labeled and unlabeled records with 20:80 ratio} \\
\# \textit {unlabeled\_records have their class field set to -1 }
\item test\_records = select\_next\_test\_fold() \\
\# \textit {Concatenate labeled and unlabeled records to get train\_records }
\item train\_records = labeled\_records + unlabeled\_records  
\item newly\_labeled = 0
\item while len(labeled\_records) $\ < $ len (train\_records):
\begin{enumerate}[5.1]
\item lp\_probability\_matrix = run\_lp(labeled\_records + unlabeled\_records)
\item model = fit\_classifier(classifier, labeled\_records)
\item labeled\_atleast\_one = False
\item for record in unlabeled\_records:
\begin{enumerate}[i.]
\item classifier\_out = model.predict\_class(record.feature\_vector)\\
\# \textit {Test for LP and classifier agreement}
\item if lp\_probability\_matrix[record.feature\_vector][classifier\_out] $\ge$ threshold:
\begin{enumerate}[a.]
\item unlabeled\_records.remove(record)
\item record.class\_label = classifier\_out \#label the record
\item labeled\_records.add(record) \#add the newly labeled record to set of labeled records
\item newly\_added += 1\\
\# \textit {Set labeled\_atleast\_one flag to True if at least one new record is labeled in current iteration of while loop }
\item labeled\_atleast\_one = True \\
\# \textit { Break the loop if no new record is labeled in current iteration of while loop }
\end{enumerate}
\end{enumerate}
\item if labeled\_atleast\_one == False:
\begin{enumerate}[5.5.1]
\item break\\
\# \textit { Compute F-measure of constructed model }
\end{enumerate}
\end{enumerate}
\item test\_records\_features = test\_records.get\_feature\_vectors()
\item test\_records\_labels = test\_records.get\_labels()
\item predicted\_labels = model.predict(test\_records\_features)
\item f-measure = compute\_fmeasure(predicted\_labels, test\_records\_labels)
\end{enumerate}

\subsection{Parallel Version}
Input: Classifier, Threshold, No\_of\_tasks \#Number of parallel processes\\
Output: F-measure
\begin{enumerate}[1.]
\item newly\_labeled = 0
\item while len(labeled\_records) $<$ len(train\_records):
\begin{enumerate}[2.1]
\item lp\_train\_records = labeled\_records + unlabeled\_records
\item lp\_probability\_matrix = []; classifier\_out = []
\item lp\_process = new\_process(target = run\_lp, args = (lp\_train\_records, lp\_probability\_matrix))
\item lp\_process.start()
\item classifier\_process = new\_process(target = fit\_classifier, args = (classifier, labeled
\_records, unlabeled\_records, classifier\_all\_out))
\item classifier\_process.start()
\item lp\_process.join()
\item classifier\_process.join()
\item atleast\_one\_labeled = False
\item chunk\_size = len(unlabeled\_records) / No\_of\_tasks
\item all\_pids = []
\item None\_initialize(labeled\_lists, No\_of\_tasks)
\item None\_initialize(unlabeled\_copies, No\_of\_tasks)
\item for i in range(len(labeled\_lists)):
\begin{enumerate}[i.]
\item start = i * chunk\_size 
\item end = (i+1) * chunk\_size
\item unlabeled\_copies = unlabeled\_records[start : end]
\item lp\_probabilities = lp\_probability\_matrix[start : end]
\item classifier\_outs = classifier\_all\_outs[start : end]
\item label\_records\_process = new\_process(func = label\_data, args = (unlabeled\_copies[i], labeled\_lists[i], lp\_probabilities, classifier\_outs, threshold))
\item label\_records\_process.start()
\item all\_pids.append(label\_records\_process)
\end{enumerate}
\item unlabeled\_records = []
\item done\_processes = []
\item while len(done\_pids) $< $ len(all\_pids):
\begin{enumerate}[i.]
\item for i in range(len(all\_pids)):\\
{\tiny .}\hspace{.40cm} if not all\_pids[i].is\_alive() and (i not in done\_pids):
\begin{enumerate}[a.]
\item done\_processes.append(i)
\item unlabeled\_records += unlabeled\_copies[i]
\item labeled\_records += labeled\_lists[i]
\end{enumerate}
\end{enumerate}
\item if atleast\_one\_labeled == False:
\begin{enumerate}[2.18.1]
\item break\\
\# \textit {Compute F-measure of constructed model}
\end{enumerate}
\end{enumerate}
\item predicted\_labels = [ ]
\item test\_records\_features = test\_records.get\_feature\_vectors()
\item test\_records\_labels = test\_records.get\_labels()
\item run\_parallel\_classifier(predicted\_labels, labeled\_records, test\_records\_features, classifier, no\_of\_tasks)
\item f-measure = compute\_fmeasure(predicted\_labels, test\_records\_labels)
\end{enumerate}

\section{Experimental Section}

In our experimental analysis, we considered 12 different datasets. The datasets along with their number of attributes (excluding the class label) and number of instances are as follows (in the format–\textit {dataset : (no of attributes, no of records })): Vowel: (10, 528), Letter: (16, 10500) , Segment: (18, 2310), Iris scale random: (4, 149), Satimage: (36, 1331), 10000 – SDSS : (7, 10000), 1000 – SDSS : (7, 1000), Glass scale random : (9, 214), Letter 1 : (16, 4500), Mfeat : (214, 2000), Pendigits : (16, 7494) and Shuttle : (9, 12770). 
In our experimental analysis, the following comparisons are made.
 
1.Serial version of our hybrid approach with  Zhu et.al[16].

2.Serial version of our hybrid approach with supervised learning classifier SVM. 

3.Parallelization of our algorithm  with our own serial implementation. 

Before performing the above comparisons,the serial version of our hybrid approach is examined on the following aspects.  
The algorithm is run for different values of threshold of the probability matrix of label propagation and percentage of initially labeled data in each of the training data set. We observed the values of percentage of increase in labeled records for each iteration, percentage of labeled data at the final iteration  and finally the training time and  F-measure.An alternative classifier Logreg is also implemented and its performance is compared with SVM based on factors like F-measure and training time.

As it can be seen from Fig 1, varying the threshold of the probability matrix of label propagation has little impact on the F-measure. Considering only label propagation increase in threshold would lead to stricter labeling resulting in increase in precision and decrease in recall. Similarly, the decrease in probability threshold results in an increase in number of unlabeled records being considered for labeling. Hence, it should increase the labeling rate of the records. But when SVM is used with label propagation precision and recall are not allowed to vary significantly because a record can be labeled only when SVM and Label Propagation agree on the class label. So, unlabeled records marked by label propagation for labeling with low confidence are discarded by output of SVM. Thus the percentage of labeled data at the end of the final iteration fluctuates very little for all the thresholds (as it is shown in next graph Fig 2). So change in thresholds, has little effect on F-measure of the model.

To see the effect of the dimension and the number of records in the dataset, In Fig 3, we plotted Training time with respect to the cube of number of records in the dataset for the best performing classifier and threshold. The axis were chosen by keeping in mind the O (dim*$N^3$) complexities of both SVM and Logreg. The graph tends to show polynomial increase in training time as N increases (instead of linear). This may be the effect of neglecting lower order terms in the complexity expression of SVM and Logreg.
\begin{figure}
\centering
\includegraphics[scale=0.5]{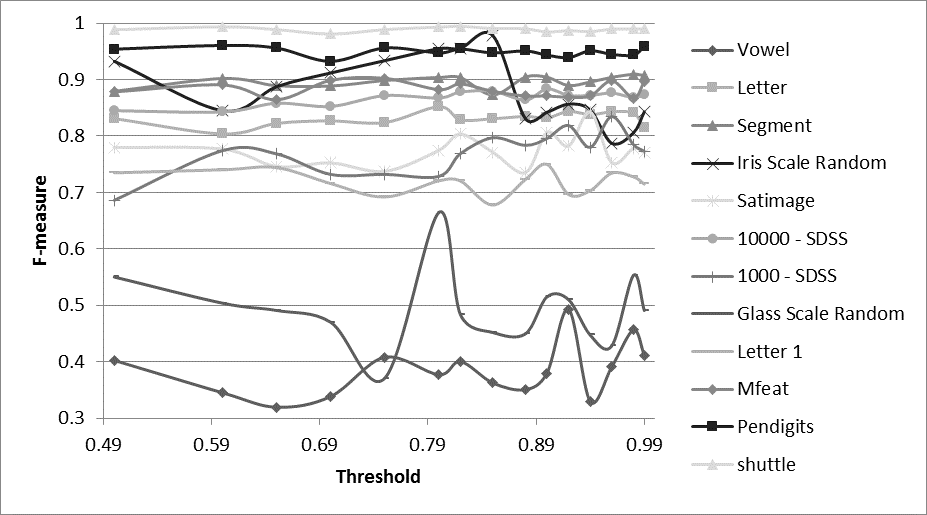}
\caption{F-measure of the datasets by varying the threshold of the probability matrix.}
%\label{fig:ex}
\end{figure}

\begin{figure}
\centering
\includegraphics[scale=0.5]{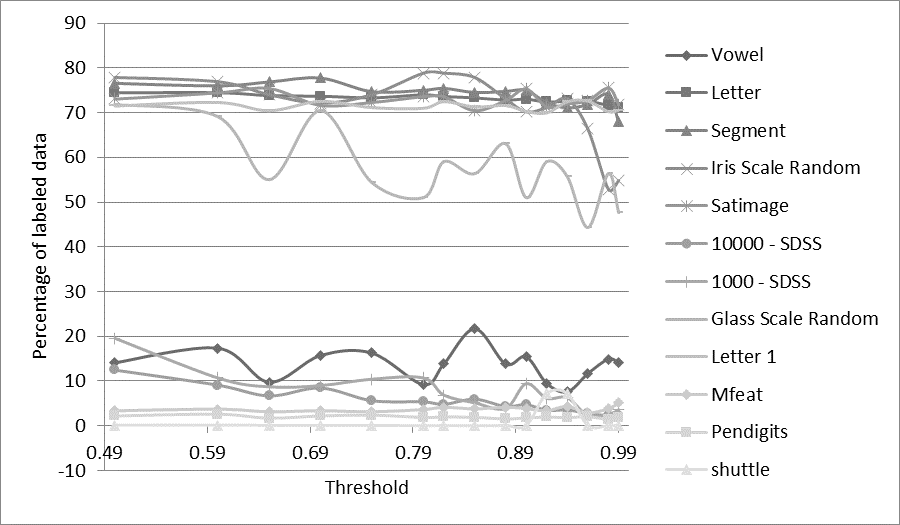}
\caption{Percentage of the labeled data by the final iteration.}
%\label{fig:ex}
\end{figure}

The analysis of Fig 1 and Fig 2 explains the following feature of the algorithm. For the datasets: 1000 records – Sloan Digital Sky survey (SDSS), 10000 records –SDSS, Mfeat, Pendigits and Shuttle,Percentage of labeled data is very low 0-20\%. But F-measures are reasonably high between 0.67 to 0.9. This shows that their high F-measure does not always require high amount of unlabeled data to be labeled. As long as the algorithm is able to label representative records in the dataset, it is likely to give good F-measure.

\begin{figure}
\centering
\includegraphics[scale=0.5]{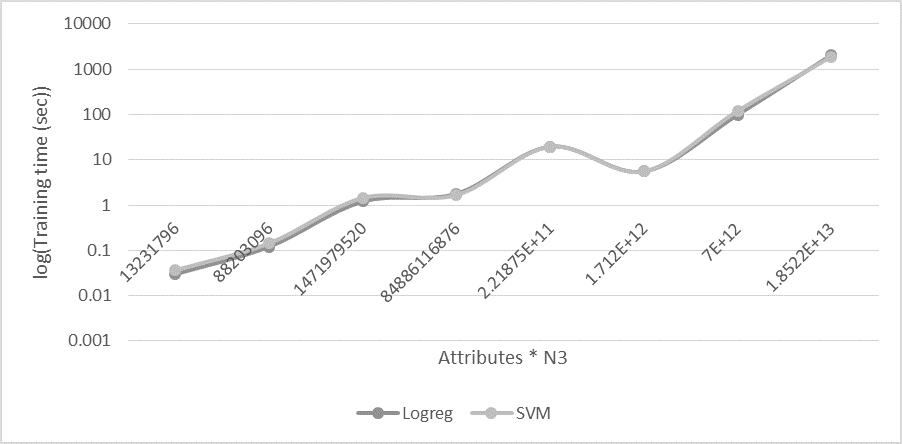}
\caption{Training time of the model by using different classifier SVM and Logreg.}
%\label{fig:ex}
\end{figure}

We observed the  percentage of increase in labeled records  for every iteration for the best choice of classifier and threshold (according to F-measure). In Fig 4, it shows that percentage of increase in labeled records decreases exponentially (note the log scale) as the iterations progress. This means not much data gets labeled as loop progresses. This is the consequence of Label propagation and SVM not agreeing on deciding the class label which is to be assigned to the unlabeled record. While labeling an unlabeled record, there is a low chance of misclassification by SVM , since it is always trained on labeled data. This means that the quality of labeling done by Label propagation decreases significantly as the  iterations progresses. This deterioration in Label Propagation’s quality has very little effect on algorithm’s overall prediction quality because of the  right  predictions done by SVM at every step while labeling the unlabeled records leading to better performance than Label propagation.

\begin{figure}
\centering
\includegraphics[scale=0.5]{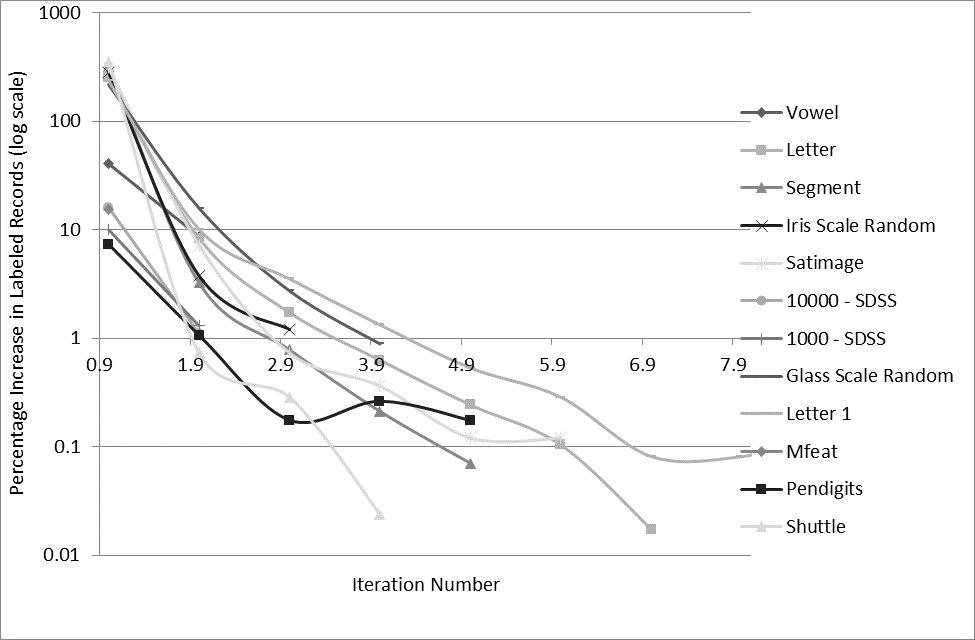}
\caption{Percentage of increase in labeled records for every iteration.}
%\label{fig:ex}
\end{figure}

The performance of the proposed approach is compared with the label propagation algorithm[16] for all the datasets and shown in Fig 5. F-measures of the proposed approach were noted for best choice of classifier and threshold. For label propagation, all the unlabeled records were labeled according to the class corresponding to highest label propagation Probability. In all the cases, the proposed approach outperforms label propagation by a large margin. This high quality of performance can be attributed use of SVM together with Label Propagation to label the unlabeled examples. No unlabeled example is labeled without the agreement of both – SVM and label propagation. This significantly reduces the pitfalls caused by label propagation increasing the prediction quality of overall approach.

The analysis of our approach(semi supervised learning) with the supervised SVM is also studied. Results in Fig 6  show that the F-measure of our approach is comparable.

Fig 7 to 12  are plotted for different percentages of initially unlabeled data for the data sets Vowel, Irisscalerandom, Satimage, 1000-SDSS, Glassscalrandom, Mfeat respectively – considering best choice of Threshold (as per the F-measure). As can be seen, F-measure of the model tends to fall as percentage of initially unlabeled data increases. This is intuitive. As the amount of labeled data in the initial data increases, the algorithm is able to learn the pattern present in a representative sample of the dataset. Thus it can successfully generalize the pattern to the test set leading to a overall increase in F-measure.

\begin{figure}
\centering
\includegraphics[scale=0.5]{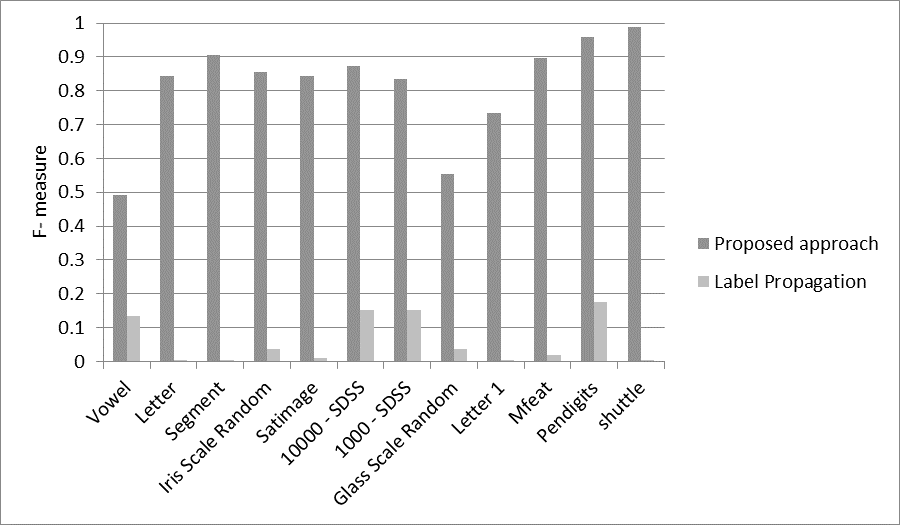}
\caption{The comparison of the proposed approach with the label propagation[16].}
%\label{fig:ex}
\end{figure}

\begin{figure}
\centering
\includegraphics[scale=0.5]{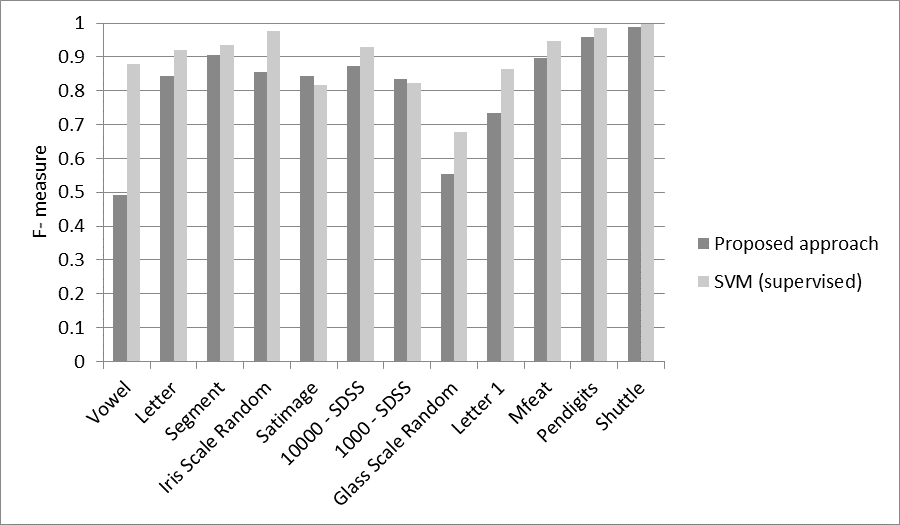}
\caption{The comparison of proposed approach with supervised one-one SVM}
%\label{fig:ex}
\end{figure}

\begin{figure}
\centering
\includegraphics[scale=0.5]{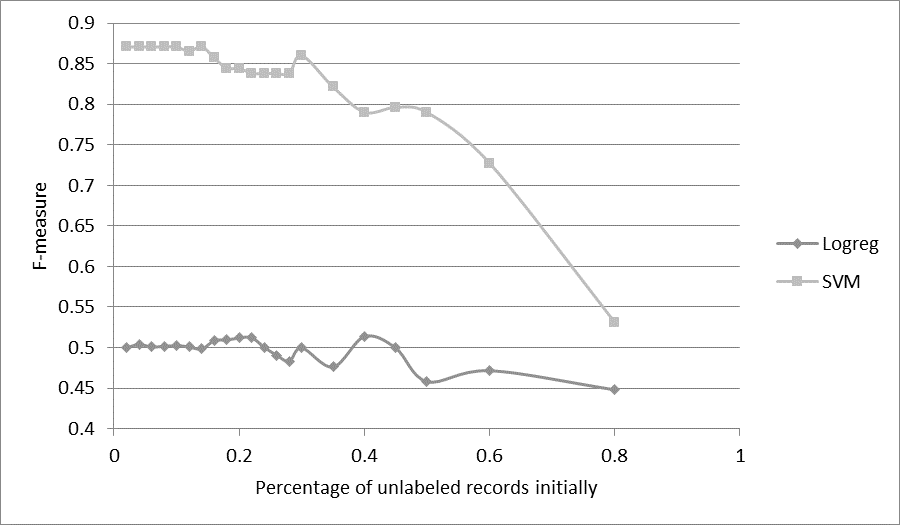}
\caption{F-measure of Vowel data set by varying the percentage of initially unlabeled records}
%\label{fig:ex}
\end{figure}

\begin{figure}
\centering
\includegraphics[scale=0.5]{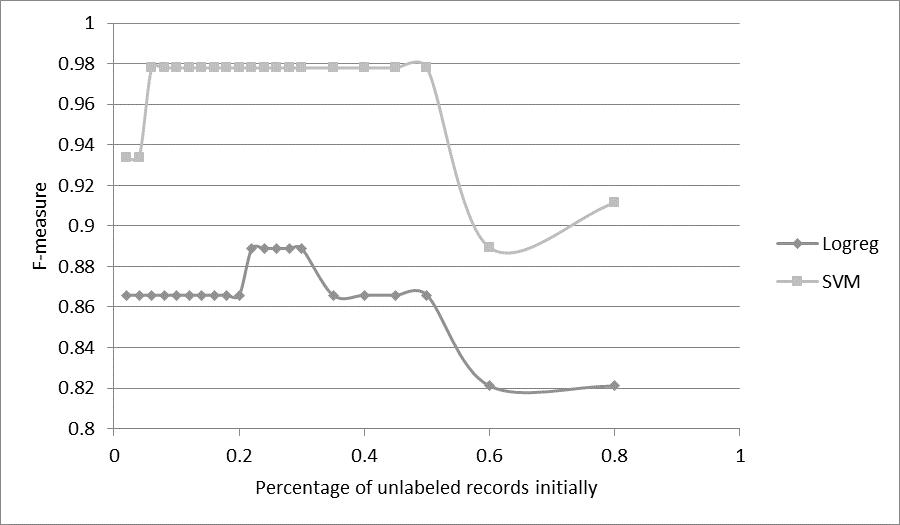}
\caption{F-measure of Irisscalerandom data set by varying the percentage of initially unlabeled records}
%\label{fig:ex}
\end{figure}

\begin{figure}
\centering
\includegraphics[scale=0.5]{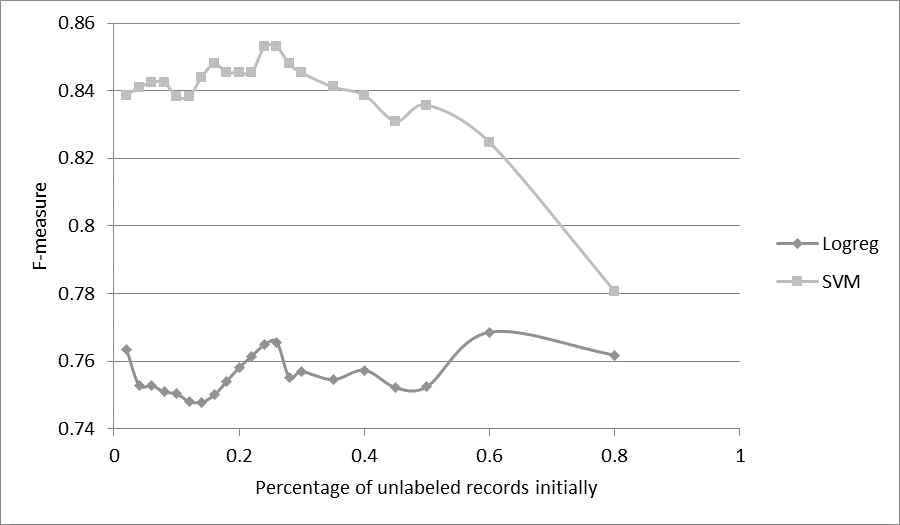}
\caption{F-measure of Satimage data set by varying the percentage of initially unlabeled records}
%\label{fig:ex}
\end{figure}

\begin{figure}
\centering
\includegraphics[scale=0.5]{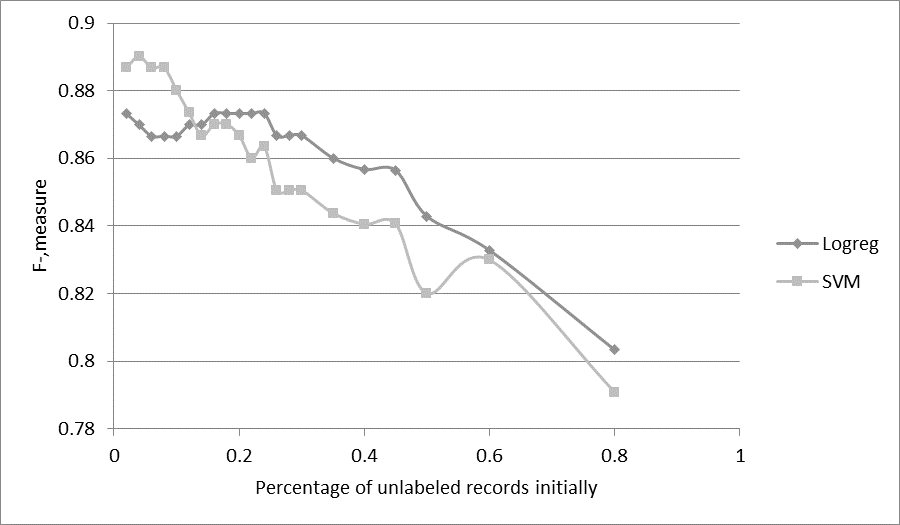}
\caption{F-measure of 1000-SDSS data set by varying the percentage of initially unlabeled records}
%\label{fig:ex}
\end{figure}

\begin{figure}
\centering
\includegraphics[scale=0.5]{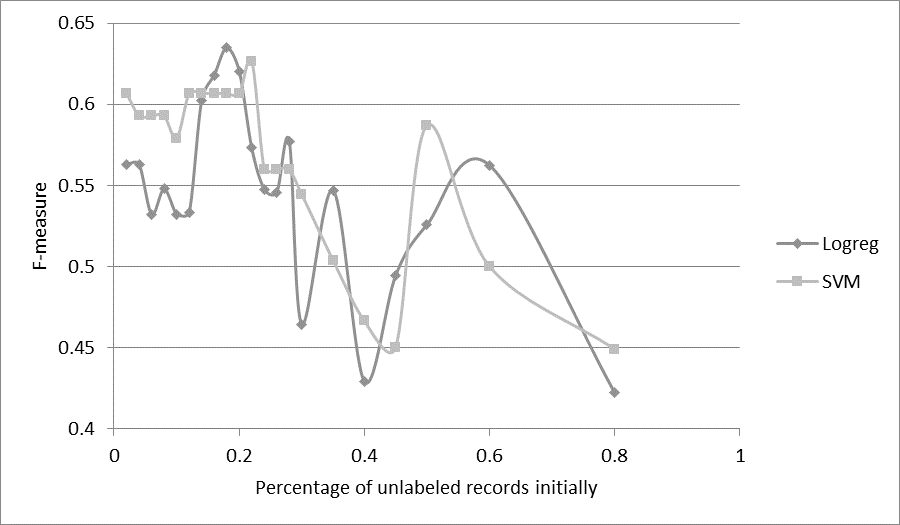}
\caption{F-measure of Glassscalrandom data set by varying the percentage of initially unlabeled records}
%\label{fig:ex}
\end{figure}

\begin{figure}
\centering
\includegraphics[scale=0.5]{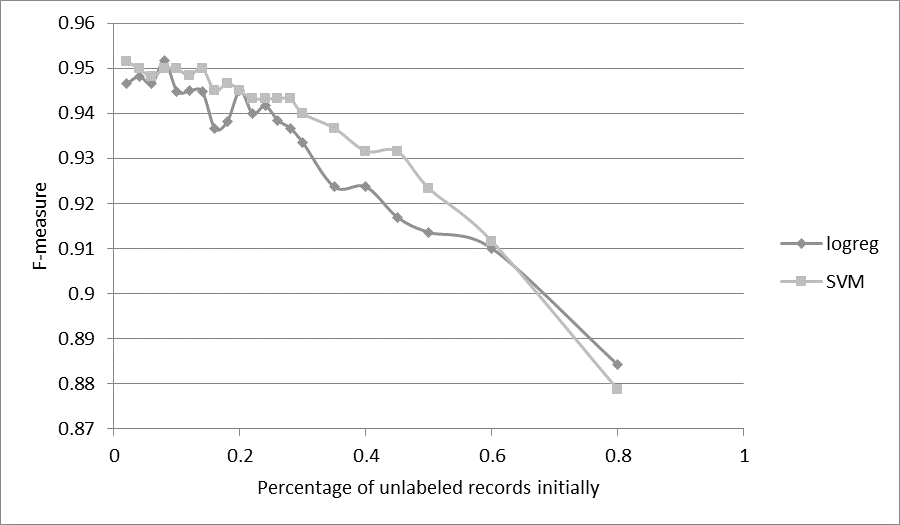}
\caption{F-measure of Mfeat data set by varying the percentage of initially unlabeled records}
%\label{fig:ex}
\end{figure}

 \begin{figure}
\centering
\includegraphics[scale=0.5]{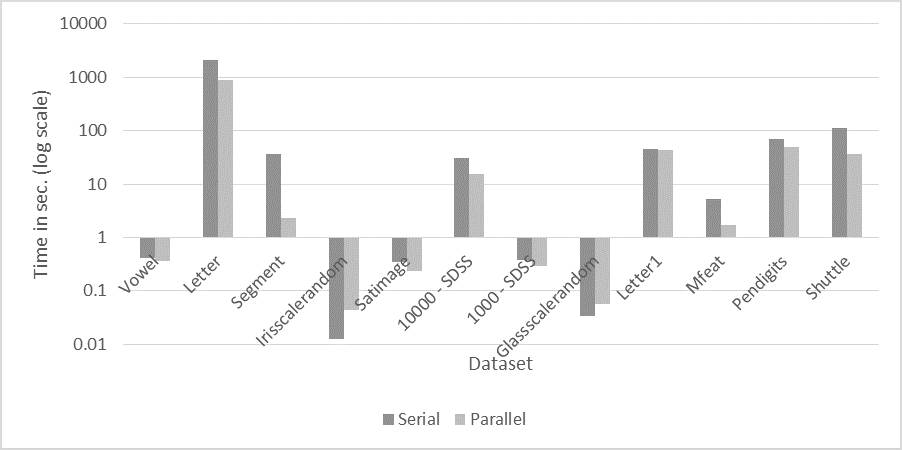}
\caption{The comparison of the proposed approach of serial and parallel versions. }
%\label{fig:ex}
\end{figure}

Finally, The parallel version of the proposed approach is also implemented. As the results can be seen from Fig 13, parallelizing the algorithm helps to improve training time of the algorithm. Two of the most expensive steps in the algorithm are training SVM and label propagation on the data. Doing these two in parallel reduces training time significantly (note that training time is in log scale).

The analysis is done for SDSS dataset for samples of different sizes.The results are shown in Fig 14. For each sample, we ran different number of parallel tasks and training time is observed. Results show that number of parallel tasks have reasonable effect on training time only when dataset size exceeds a certain threshold (around 60000 records in this case). Further, for each dataset, there is an optimum number of parallel tasks which yields minimum training time. If number of parallel tasks is above this optimum level, the cost of maintaining these parallel tasks exceeds the gain earned by parallel computation. On the other hand, if number of parallel tasks is set to a value less than optimum level, system resources are poorly utilized. So it is necessary to set number of parallel tasks to optimum value for maximum benefit.

\begin{figure}
\centering
\includegraphics[scale=0.5]{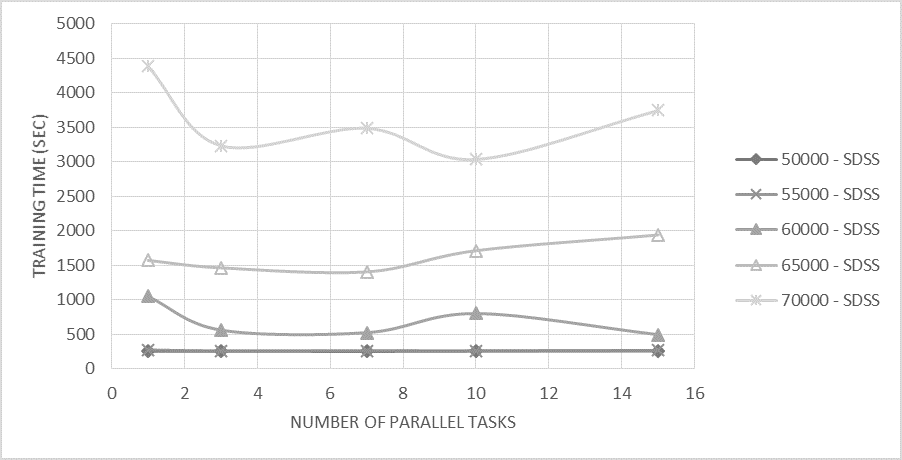}
\caption{The number of parallel tasks w.r.t. training time}
%\label{fig:ex}
\end{figure}

The proposed approach is tested on skewed datasets also. The proportion of class labels in the skewed dataset is at least 1:8. F-measure of 10000-SDSS drops by approximately 0.1 when it has skewed class proportions. But skewed version of Shuttle shows exceptional behavior. Its F-measure remains almost the same. This shows that skewness of the data has little or no effect on F-measure of algorithm. It can be inferred that the distribution of the data plays a major role in performance of semi-supervised algorithm. The results are shown in Fig 15.

\begin{figure}
\centering
\includegraphics[scale=0.5]{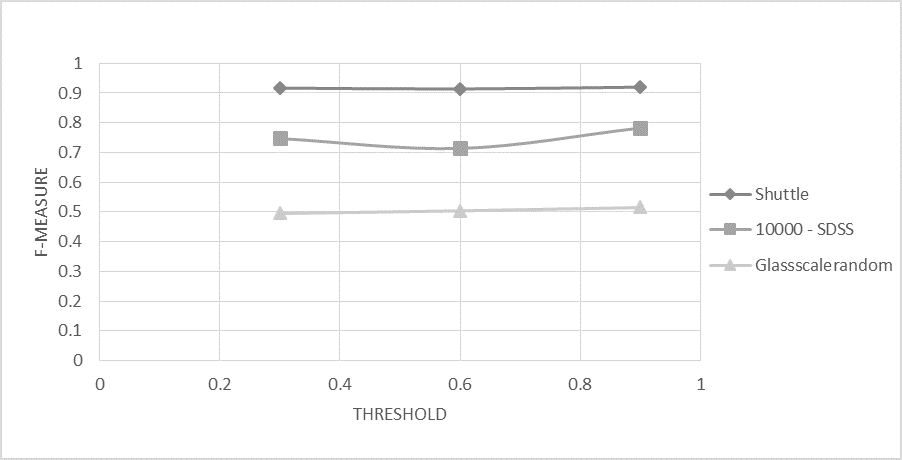}
\caption{The analysis of F-measure of skewed data sets.}
%\label{fig:ex}
\end{figure}

\section{Conclusion and Future work}
The proposed approach uses SVM along with Label Propagation algorithm to yield a very high overall prediction quality. It can use a small amount of labeled data along with a large quantity of unlabeled data to yield a high F-measure on test data. It has a very small margin for error since it labels the unlabeled data using consent of both - SVM and Label Propagation. On testing both the algorithms on 12 different datasets we can conclude that the proposed approach performs much better than label propagation[16] alone. It yields F-measure values which are almost twice as compared to Label Propagation. Further, we designed the parallel version of the approach and were able to decrease the training time significantly. In future, the parallel algorithm can be further enhanced to yield linear or super linear scale up. Further research on the role of supervised algorithms in the field of semi supervised learning could be beneficial.

\medskip

\noindent

\subsubsection*{Acknowledgments.} We are thankful for the support provided by the Department of Computer Science and Informations Systems, BITS, Pilani, K.K. Birla Goa Campus to carry out the experimental analysis.

\end{document}